# ASR Bundestag: A Large-Scale political debate dataset in German

Johannes Wirth and René Peinl

University of Applied Sciences Hof, Alfons-Goppel-Platz 1, 95028 Hof, Germany

**Abstract.** We present ASR Bundestag, a dataset for automatic speech recognition in German, consisting of 610 hours of aligned audio-transcript pairs for supervised training as well as 1,038 hours of unlabeled audio snippets for self-supervised learning, based on raw audio data and transcriptions from plenary sessions and committee meetings of the German parliament. In addition, we discuss utilized approaches for the automated creation of speech datasets and assess the quality of the resulting dataset based on evaluations and finetuning of a pre-trained state of the art model. We make the dataset publicly available, including all subsets.

## 1 Introduction

As current neural networks grow in terms of the number of their parameters, more and more data are needed to train robust and well-performing models [1]. Especially in the field of automatic speech recognition (ASR), current model architectures like Wav2Vec 2.0 [2, p. 2] are based on self-supervised training with large amounts of unlabeled data. However, finetuning is usually still necessary and done in a supervised fashion that requires significant amounts of labeled data, especially in use cases involving domain-specific vocabulary or in spontaneously spoken sentences, which often lack correct grammar. In their analysis, Wirth and Peinl [3] show that models trained with more labeled data like Conformer-CTC and Citrinet outperform Wav2vec 2.0, which was trained on a large unlabeled dataset and finetuned with only few labeled data. HuBERT X-Large is currently one of the largest ASR models with one billion parameters [4] and seems still undertrained, which leads [5] to further scale their training data to 96k hours, although their large model has only 316 million parameters. Chan et al. [6] show with their SpeechStew strategy, that using a large number of transcribed speech from different domains together for supervised training leads to a well generalizing model that reaches word error rates (WERs) close to the specialized state-of-the-art (SOTA) models for a number of datasets. While datasets for speech recognition in English, both read aloud and spontaneously spoken, exist in very large quantities, e.g. SPGISpeech [7] with 5,000 hours or Gigaspeech [8] with even 10,000 hours of transcribed audio, relatively few datasets exist in German, which is also evident when comparing the performance of trained SOTA for these languages.

We present ASR Bundestag, a dataset for speech recognition consisting of political debates of the German Bundestag (parliament) with subsets for supervised and unsupervised training of speech recognition models in German, which reflects large amounts of spontaneous speech as well as political terms and phrases and can improve the performance of models for the German language.

This work is structured as follows: First, an overview of currently freely available speech datasets in German is given, followed by a description of the generation process of the dataset presented in this paper. Afterwards, the properties and subsets of the dataset are described in more detail and the results of finetuning experiments in which it was utilized are presented. Finally, these results are interpreted, and all previous points are summarized in a conclusion along with an outlook.

## 2     German datasets for speech recognition

Since German is considered a high-resource language, there are several datasets available that can be used for speech recognition. Some of them are specific to ASR and created with ASR in mind, e.g., M- AILABS [9] or Mozilla Common Voice (MCV) [10], others can be taken from speech synthesis efforts like HUI [11] and Thorsten Voice neutral [12]. Some datasets were created for linguistic research like BAS ALC [13] and SC10 [14] but can still be used for ASR training.

Furthermore, available datasets can be categorized into read speech and spontaneous speech. The latter can further be divided into conversations and talks. Talks can be described as continuous switching between read aloud and spontaneous speech as they may be scripted to a certain degree, but usually are not directly read and contain at least a proportion of spontaneous parts. Examples for read speech include Voxforge [15] and Spoken Wikipedia Corpus [16]. Spontaneous speech datasets include German TED Talks [17] and VoxPopuli [18]. However, conversational datasets are still scarce and limited in the amount of data. The 25 hour Hempel dataset [19] and the recently published 20h subset by Datatang for Interspeech 2022 [20] are a few noteworthy exceptions. One challenge for conversational speech is how to cope with colloquial expressions and swallowed syllables. Mapping speech like this to a standard German text as it is done e.g. by [20] is mixing up two steps into a single one, namely a phonetic transcription of audio into characters and then from dialect or colloquial speech into standard German text. Using such data for training can improve the performance of ASR models for this special dataset, but at the same time also decrease overall performance across a large number of datasets. Ideally, such datasets would come with two separate transcriptions, one with standard German text and one with a phonetically correct transcript.

Some datasets are also concerned with technical aspects of the recorded speech like Tuda-De [21] and Verbmobil 2 [22]. They provide speech recorded with different microphones at the same time and mobile phones respectively, so that ASR systems can specifically be trained or evaluated in this regard.

The largest dataset with over 3,000 hours of German speech is MLS [23], which is provided by Facebook. Like M-AILABS and Voxforge, it is based on LibriVox [24].

However, its quality is rather inferior due to missing normalization of numbers and abbreviations in German [3].

Finally, gender bias can be tested with datasets like SL100 [25] and Hempel [19] that contain an equal amount of female and male speakers. Since this is not the case for the large datasets used for training (especially for MCV), there are reports about gender bias, which puts women at a disadvantage compared to men [26]. Surprisingly, in their evaluation Garnerin et al. found this gender bias even if women are not underrepresented in training data.

Table 1. Datasets for ASR and TTS in German language compared to Bundestag.

| Dataset | Description | Hours |
| --- | --- | --- |
| BAS ALC | 162 speakers, sober and intoxicated | 95 |
| Hempel | 3920 speakers, across states, 47% male, spontaneous speech | 25 |
| BAS SI100 | 101 speaker, 50% male, read text from newspapers | 32 |
| SC10 | 70 speakers, read and non-prompted speech | 12 |
| VerbMobil 2 | ASR, multi-lingual, 401 speakers, conversational | 22 |
| German TED-Talks | ASR, German, 71 speakers, spontaneous speech | 16 |
| MCV 7.0 | ASR, multi-lingual,15,620 speakers, read speech | 1,062 |
| TUDA-De | ASR, German, 147 speakers, read speech | 127 |
| SWC | ASR, multi-lingual, 363 speakers, read speech | 285 |
| M-AILabs | ASR, multi-lingual, based on LibriVox and others | 237 |
| MLS | ASR, multi- lingual, 244 speakers, based on LibriVox | 3,287 |
| Voxforge | ASR, multi- lingual, based on LibriVox | 35 |
| HUI | TTS, German, 122 speakers, based on LibriVox | 326 |
| Thorsten | TTS, German, 1 speaker, read speech, very clear | 23 |
| VoxPopuli | ASR, multi-lingual, political debates from EU parliament | 268 |
| Bundestag Clean | | 610 |
| Bundestag Dirty | ASR, German, political debates from German Bundestag | 766 |
| Bundestag Unlabeled | | 1,038 |

Besides MLS, there are no large scale ASR datasets in German language like SPGISpeech [7] with 5,000 hours or Gigaspeech [8] with 10,000 hours of transcribed audio. And even for MLS [23] with its dubious quality in German, there is far more data for English than for German (71,506 compared to 3,287 hours). Therefore, it is not surprising that the best models achieve word error rates (WER) of 1.4% and 2.5% on LibriSpeech test clean and other datasets [27], whereas WERs on M-AILABS and Tuda-De, which are among the most cited datasets for German ASR, are still comparably high with 4.3% and 5.8% [3].

With the contribution of a new high-quality dataset for speech recognition called ASR Bundestag, this paper tries to contribute to progress in ASR for German language.

# 3    Data Acquisition and Processing

For the creation of the dataset, recordings, and transcripts of sessions of the German Bundestag were first accumulated, processed into short audio snippets as well as normalized transcripts and subsequently aligned into short audio transcript pairs using two distinct processing methods.

## 3.1    Acquirement of Data

The media library of the German Bundestag [28] provides recordings of plenary sessions and committee meetings to the public and permits their use for non-commercial purposes [29]. 331 recorded and transcribed sessions of the German Bundestag consisting of 1,200 hours of plenary session recordings as well as committee meeting recordings served as a basis for the dataset and were processed using the methods described hereafter.

## 3.2    Data Preprocessing

Since all audio was recorded at a sampling rate of 44.1 kHz stereo and stored in mp3 format, the sampling rate was converted to 16 kHz and the number of channels from stereo to mono, as well as the file format to WAV to ensure that the audio data could be used as input for considered speech recognition models.

Additionally, transcripts were normalized in terms of capitalization, abbreviations, and numbers using an internally developed normalizer. Of 331 available sessions, 320 were utilized for dataset generation, whereas the remaining eleven were excluded due to missing transcripts.

## 3.3    Alignment Methods

The previously accumulated audio transcript pairs, consisting of very long recordings (3.9 hours on average), were further processed in a subsequent step using two distinct methods for the alignment of audio and transcript snippets.

### 3.3.1    Splitting and Alignment based on Timestamps and CTC

Since the transcripts of all used recordings include timestamps along with additional metadata (change of speaker, applause, etc.), audio recordings were initially split using these and transcribed with a Citrinet model, pre-trained on MCV7.0, MLS and VoxPopuli, provided through the NVIDIA NeMo toolkit [30] in order to force alignment of transcript snippets and transcript predictions. A comparison of the original Bundestag transcripts to the transcript predictions of the ASR model revealed that the annotated transcripts differ greatly from what was actually said. This was partly due to corrections of repeated words or sentence structure, but mainly because the transcript timestamps vary significantly in terms of accuracy.

For this reason, the approach of timestamp-based splitting was extended with temporal tolerances (an additional five seconds at the beginning and end of each audio snippet) to increase the likelihood that transcripts are contained entirely within the audio recordings split with temporal tolerance. To subsequently narrow down the previously established temporal tolerances again, transcript predictions using a model with a CTC decoding scheme are used and CTC outputs are matched. In this process, the time step at which the greedy output token of the ASR model matches the beginning of the corresponding transcript is determined and the previous time steps of the audio recording are removed. Analogously, the same technique is applied to remove additional audio after the time step of the last token occurring in the transcript.

Although the described forced alignment method can generate precise alignments between provided transcripts and audio snippets, it is not suitable for the creation of an ASR dataset as transcripts differ too much from what is said, preventing exact alignments between speech and transcripts. This is mostly due to spontaneous speech, including slips of the tongue, stutters and incorrect sentence structure, which are removed or corrected in transcripts. These circumstances complicate the alignment of transcript and audio recordings.

3.3.2    Splitting and Alignment based on Speaker Diarization and Silence Intervals

To obtain audio transcripts that match utterances and sentences, which were actually spoken in audio recordings as precisely as possible, long session recordings are first split using output predictions of a neural network for speaker diarization [31] (pyannote.audio 2.1.1) and segmented outputs of this diarization and splitting process longer than 30 seconds are again split based on silence thresholds. Splits are generated in a range from +60 dB to -60 dB, and the ones with the longest average duration are kept for further processing. In a subsequent step, transcript predictions are generated for the resulting audio snippets using a pre-trained speech recognition model (Citrinet as mentioned in 3.3.1) and utilized as references for alignment.

In a forced alignment process, each transcript prediction is matched to a part of the transcript with a minimum character error rate (CER). This is accomplished by iterating over the entire transcript and determining a string with the length of the transcript prediction and an additional five words before and after it, for which the CER is minimal. The transcript start and end are then specified using the CTC-based narrowing described in 3.3.1, whereas in this case the transcript candidate is fitted rather than the audio recording.

Instead of the matching process described here, CTC-segmentation as described in [32] could have been used, but consistent transcription inaccuracies, especially with spontaneous speech in [3], have been found with the ASR model used. This is indicated by "hallucinations", words that are at the beginning of transcript predictions but are not said in the transcribed audio. This is a symptom of pre-training with consistently erroneous data. These hallucinations can lead to heavy penalties during the alignment process. Additionally, we chose not to use utilize this method since Bundestag data primarily contains spontaneous speech, whereas CTC-segmentation is best suited for read out texts.

## 4 The German Bundestag dataset

In the following section, the metadata of the datasets generated by the previously described alignment methods and subsets filtered based on further metrics are described in detail.

### 4.1 Alignment based on Timestamps and CTC

Due to the limitations mentioned before, the dataset generated using timestamps and CTC alignment cannot be used directly for the efficient training of speech recognition models in German and no pre-training or finetuning experiments have been conducted with it. Nevertheless, it is made available to the public as it represents contextually accurate audio transcript pairs for other potential applications.

### 4.2 Alignment based on Silence Intervals and Speaker Diarization

Based on the original 1,200 hours of raw data, a dataset consisting of 530,419 audio transcript pairs with a total length of 1,038 hours was created. Since the alignment process aimed to minimize CER between transcript snippets and transcript predictions, audio transcript pairs also contain suboptimal alignments, which were filtered out by setting a maximum deviation threshold and further algorithms. Experiments and sampling have shown that a maximum CER of no more than 15% between transcript and reference (transcript prediction) result in generally high-quality alignments.

However, words often occur at the beginnings and ends of transcript snippets that are not said in the associated audio snippets. Similar issues were identified when examining the Voxpopuli dataset [18], which was created by conducting similar, threshold-based filtering (using 20% CER). To minimize this effect, two additional conditions were set for audio transcript snippets to be defined as well-aligned: First and last word matching between transcript and transcript prediction are the same, or transcript and transcript prediction contain the same number of words.

The described filtering of the audio transcript snippets resulted in a split of the overall dataset into parts of varying alignment quality, which are referred to as "clean", "dirty" and "unlabeled" hereafter.

**Clean**:
As previously described, audio transcript pairs are placed in the clean category if a transcript snippet is found that is off by less than 15% CER from a transcript reference and either the transcript and transcript reference contain the same number of words or the first and last word match exactly. In addition, transcripts with less than five words were excluded because longer recordings yield higher confidence with the same relative transcript deviation and very short audio transcript snippets should not be included in this subset. This exclusion is also reflected in the distribution of audio durations of this subset (Fig. 1), compared to the distributions of the supersets dirty (Fig. 2) and unlabeled (Fig. 3) where no such restrictions were established. Initial fine-tuning experiments using the data processed by this approach were successful, but revealed a total of ~2% outliers, which were subsequently investigated, excluded, and included in the "Unlabeled" superset. Lastly, alignments in which the first and last word did not match exactly were evaluated and semi-automatically corrected. Audio transcript pairs that satisfy the previously established restrictions amount to 58.8% of the total data (~610 hours). The dataset was randomly split into training, validation, and test splits (90/5/5), with similar average audio sample duration (± 0.01 seconds) and similar sample duration distribution added as conditions.

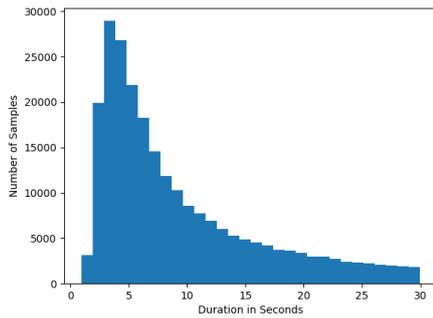

**Fig 1.** Distribution of audio durations for ASR Bundestag Clean.

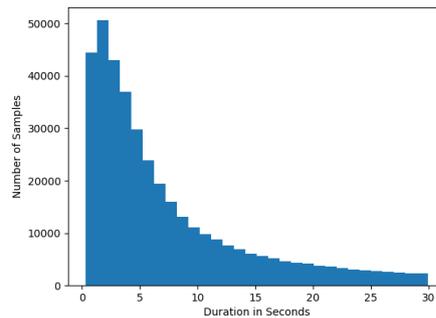

**Fig 2.** Distribution of audio durations for ASR Bundestag Dirty.

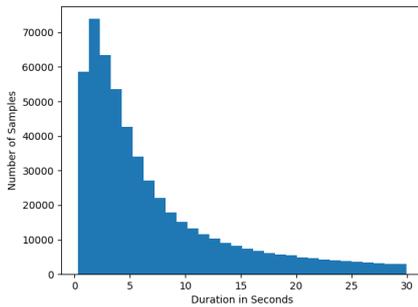

**Fig 3.** Distribution of audio durations for ASR Bundestag Unlabeled.

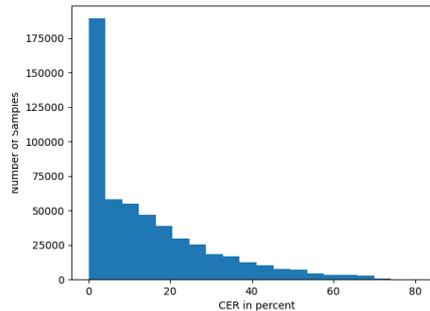

**Fig 4.** Distribution of matchings between audio and transcript snippets based on CER.

**Dirty**:

For this subset, the degree of deviation between transcript and transcript prediction was increased to 20% CER and no lower limit was set with regard to transcript length. This subset is still conditionally suitable for use as a training dataset for automatic speech recognition models. The CER threshold was set according to the filtering process as described in the processing segment of Voxpopuli (transcribed speech), in order to perform a comparison between both datasets in the future. This part, with the constraints lifted, forms a superset of the "clean" subset and additional subset of the total data. Audio transcript pairs in this set amount to ~73.8% of the total data (~766 hours).

**Unlabeled**:

The all-encompassing superset, which comprises the dirty and thus also clean subset, is called "unlabeled", since it also includes all automatically created audio-transcript pairs, even for those of which no alignment of transcripts and references could be determined with a CER of less than 20% (the full distribution is shown in Fig. 4). The average probability that what is said in an audio snippet matches the corresponding transcript closely is too low in this dataset to train a robust ASR model in a supervised manner. However, it can be used for semi-supervised and unsupervised training. The total duration of all audio snippets in this set totals 1,038 hours.

## 5   Speech recognition results

ASR Bundestag clean was used to finetune the Citrinet model previously used for the creation of transcript predictions in the alignment process of dataset creation as well as a Conformer-CTC model pre-trained on the same training datasets. The basic configuration used for pre-training was retained, with the following parameters modified:

- CosineAnnealing was kept as scheduler, only the maximum learning rate was decreased from $5x10^{-2}$ to $1x10^{-4}$.
- Warmup was removed.
- Maximum input audio duration was increased from 20 to 30 seconds, for adaptation to the training dataset.
- Batch size was set to 8 to fit the GPU used.
- Training duration was fixed to 30 epochs (50 epochs for Conformer-CTC).

The processes were carried out utilizing a NVIDIA RTX A6000 GPU.

**Table 2: Performance results and metadata for ASR models finetuned on ASR Bundestag clean.**

| Model Architecture | Citrinet | Conformer-CTC |
|---|---|---|
| Pre-training datasets | MCV7.0, MLS, VoxPopuli | |
| # of Epochs | 30 | |
| Finetuning duration | 87h | 76h |
| Pretraining WER mean % | 10.68 | 13.14 |
| Pretraining WER median % | 10.00 | 11.90 |
| Finetuning WER mean % | 8.48 | 9.41 |
| Finetuning WER median % | 7.14 | 8.00 |
| Absolute improvement % | 2.20 | 3.73 |
| Relative improvement % | 20.60 | 28.39 |

As shown in Table 2, finetuning improved the performance of both models in terms of WER relatively by over 20%. However, median WERs are significantly below mean values, which may be an indication of few transcript predictions with extremely high WER but overall better model finetuning than indicated by the mean values.

Both fine-tuned models show a significant relative improvement of the word error rate on the test data of the Bundestag dataset, although both have already been pre-trained with a fairly high amount of data. These improvements demonstrate consistently accurate alignments between audio and transcript snippets of the Bundestag dataset.

It is also likely that further training will lead to even greater improvements as loss values continued to converge until the training was finished. However, hyperparameter optimization should be conducted before further finetuning experiments.

In-depth error evaluation revealed that ~30% of previously incorrectly transcribed words were corrected by finetuning. It was also observed that years are still transcribed as integers rather than years ("one thousand nine hundred ninety" < > "nineteen hundred ninety"). This is due to the consistently incorrect normalization of years within the Voxpopuli dataset the model was pre-trained with. While years have been correctly normalized within ASR Bundestag transcripts, the finetuning process has not yet been sufficient to correct the model to this extent.

# 6   Discussion

The description of the dataset creation process shows that it needs a major effort to create a high-quality ASR dataset. A more sophisticated filtering mechanism was necessary to distinguish between reference transcripts that seem very close to what is actually said and others that are a freer variation of the actually spoken word. Despite all iterations and enhancements that were conducted after identifying potential problems through manual quality assurance steps, there remain some doubts about the quality of the provided transcript audio alignments due to the significant difference between median and mean WER on the clean dataset. This needs to be further analyzed in future research. However, the significant improvement in WER after finetuning show that the overall quality of the dataset is nevertheless very good.

The limits regarding CER that were set during the alignment process might limit the usage for further training ASR models, since complex terms that existing models struggle with the most, might be excluded from the clean subset of the ASR Bundestag dataset. However, we saw no other feasible way to guarantee the high quality of the dataset without a massive and therefore not justifiable manual quality assurance process. A cruder automatic alignment process could include more challenging vocabulary but would significantly reduce the confidence of an exact match between audio and transcript snippets.

In addition, the impact of using ASR Bundestag clean for pre-training tasks has not yet been investigated. Since the maximum length of audio samples in ASR Bundestag is 50% higher than the data used to pre-train the fine-tuned model, loss convergence can be negatively affected when training model architectures, which are uninitialized.

Furthermore, no cross-evaluation with other test datasets has been performed yet. In particular, a comparison between Bundestag ASR and Voxpopuli would be of interest, since both have very similar characteristics regarding vocabulary, recording quality and spontaneous speech.

## 7   Related work

Besides the dataset papers referenced in section two, there are a number of other research works that deal with ASR. New architectures for speech recognition are mostly proposed and evaluated together with English datasets. Recently, the mixture-of-experts architecture that was successfully applied to large language models has also been evaluated for ASR. SpeechMoE [33] achieves a 7% to 23% relative error reduction compared to a compute-matched dense baseline on four different datasets including AIShell. SpeechMoE 2 [34] further increases the accuracy of the model with an improved routing that especially pays off for the Sichuan dialect in the evaluation of different Chinese accents.

In several other articles, more efficient versions of existing architectures have been proposed. Tiny transducer [35] is an effort to use small scale transducer models for speech recognition on edge devices. The model with 1.6 M parameters outperforms the hybrid TDNN baseline by more than 3% absolute reducing the WER from 21.5% to 18.1% on a challenging noisy in-car dataset. The Q-ASR model [37] uses quantized versions of QuartzNet and Jasper with 8 bit instead of 32 bit that performs nearly as good as the baseline, although it is only one quarter as large as the baseline.

Toolkits like SpeechBrain [38] and NeMo [39] also try to improve the accuracy of speech recognition results. SpeechBrain demonstrates in the paper several new state-of-the-art performances with existing architectures, especially on the TIMIT dataset as well as Common Voice in French and Italian. NeMo uses an own Russian dataset based on LibriVox with 98 hours of speech. Using this dataset in addition to MCV allows them to increase the accuracy on the test data from 12.6% to 9.7% compared to the baseline with MCV only.

A similar approach was taken by Bermuth et al. in the Scribosermo effort [40]. They trained a QuartzNet model on 37 German datasets in order to achieve state-of-the-art WER on the German Common Voice subset at that point in time.

Sinha and Siegert improve speech recognition for German voice assistant conversations [42]. They use the Voice Assistant Conversation in the wild (VACW) and Voice Assistant Conversation Corpus (VACC) datasets and use the ROVER algorithm to combine the transcription proposals from three different cloud-based ASR-engines (Google, IBM, Wit.ai) into a single high quality transcript. For VACC they achieve a reduction in WER from 4% of the best single ASR-engine to 3.3% for the ROVER algorithm. They were not able to achieve an improvement regarding WER for the VACW dataset.

Lopez et al. recently did an evaluation of conversational speech in Dutch, English and German [43]. They found that conversational speech has several challenges for speech recognition, namely fill words like "um" and "hm", reductions, which are especially relevant in German like 'n instead of "ein" or "ham" instead of "haben" as well as truncated words, e.g., "nich" instead of "nicht".

# 8    Conclusion and outlook

In this work we presented ASR Bundestag, a large-scale dataset for speech recognition in German, containing spoken political terms and phrases by a large number of speakers, divided into multiple subsets in varying qualities of audio-transcript alignment for supervised and self-supervised training. With a length of 610 hours, the clean subset represents one of the largest labeled datasets for speech recognition in German as well as the largest focused on political debates. In addition, the entirety of generated audio transcript pairs, based on two different segmentation processes, are published, and can be accessed at https://opendata.iisys.de/. We provide baseline evaluations using pre-trained SOTA model architecture for speech recognition in German along with results of finetuning experiments. The positive results of these experiments demonstrate the high precision of the described alignment approach and the quality of the dataset, therefore the objective of expanding high quality labeled speech data in German was accomplished.

In future work, the dataset quality will be further assessed through cross evaluations and pre-training of further architectures. Quality differences between ASR Bundestag and Voxpopuli will be reviewed. There are further plans for systematically studying the impacts of finetuning an ASR model with the ASR Bundestag dataset and evaluating its performance on other datasets, since more training can mean either better overall performance or lead to stronger specialization and therefore decreasing performance on other datasets.

# References

[1]   J. Kaplan *et al.*, "Scaling laws for neural language models," *arXiv preprint arXiv:2001.08361*, 2020.
[2]   A. Baevski, H. Zhou, A. Mohamed, and M. Auli, "wav2vec 2.0: A Framework for Self-Supervised Learning of Speech Representations," *arXiv:2006.11477*, 2020.
[3]   J. Wirth and R. Peinl, "Automatic Speech Recognition in German: A Detailed Error Analysis," in *2022 IEEE International Conference on Omni-layer Intelligent Systems (COINS)*, 2022, pp. 1–8.
[4]   W.-N. Hsu, B. Bolte, Y.-H. H. Tsai, K. Lakhotia, R. Salakhutdinov, and A. Mohamed, "Hubert: Self-supervised speech representation learning by masked prediction of hidden units," *IEEE/ACM Transactions on Audio, Speech, and Language Processing*, vol. 29, pp. 3451–3460, 2021.
[5]   S. Chen *et al.*, "Wavlm: Large-scale self-supervised pre-training for full stack speech processing," *arXiv preprint arXiv:2110.13900*, 2021.
[6]   W. Chan, D. Park, C. Lee, Y. Zhang, Q. Le, and M. Norouzi, "Speechstew: Simply mix all available speech recognition data to train one large neural network," *arXiv preprint arXiv:2104.02133*, 2021.

[7] P. K. O'Neill *et al.*, "SPGISpeech: 5,000 hours of transcribed financial audio for fully formatted end-to-end speech recognition," *arXiv preprint arXiv:2104.02014*, 2021.

[8] G. Chen *et al.*, "Gigaspeech: An evolving, multi-domain asr corpus with 10,000 hours of transcribed audio," *arXiv preprint arXiv:2106.06909*, 2021.

[9] I. Solak, "The M-AILABS speech dataset," 2019. https://www.caito.de/2019/01/the-m-ailabs-speech-dataset/

[10] R. Ardila *et al.*, "Common Voice: A Massively-Multilingual Speech Corpus," in *Proceedings of the 12th Language Resources and Evaluation Conference (pp. 4218-4222)*, Mar. 2020.

[11] P. Puchtler, J. Wirth, and R. Peinl, "HUI-Audio-Corpus-German: A high quality TTS dataset," in *44th German Conference on Artificial Intelligence (KI2021)*, Berlin, Germany, 2021.

[12] T. Müller, "Thorsten Open German Voice Dataset," Mar. 25, 2021. https://github.com/thorstenMueller/deep-learning-german-tts (accessed Mar. 26, 2021).

[13] Bavarian Archive for Speech Signals, *BAS Alcohol Language Corpus*. http://hdl.handle.net/11022/1009-0000-0001-88E5-3. 2016.

[14] V. Mapelli, *Strange Corpus 10 - SC10 ('Accents II') (ELRA-S0114), ELRA (via CLARIN VLO), ISLRN 024-991-750-952-3*. 2004.

[15] K. MacLean, *Voxforge*. [Online]. Available: http://www.voxforge.org

[16] T. Baumann, A. Köhn, and F. Hennig, "The Spoken Wikipedia Corpus collection: Harvesting, alignment and an application to hyperlistening," *Language Resources and Evaluation*, vol. 53, no. 2, pp. 303–329, 2019.

[17] F. Hernandez, V. Nguyen, S. Ghannay, N. Tomashenko, and Y. Esteve, "TED-LIUM 3: twice as much data and corpus repartition for experiments on speaker adaptation," in *Intl. Conf. on Speech and Computer*, 2018, pp. 198–208.

[18] C. Wang *et al.*, "VoxPopuli: A Large-Scale Multilingual Speech Corpus for Representation Learning, Semi-Supervised Learning and Interpretation," in *11th Intl Joint Conf. on Natural Language Processing*, 2021, pp. 993–1003.

[19] C. Draxler and F. Schiel, "Three new corpora at the Bavarian Archive for Speech Signals-and a first step towards distributed web-based recording," 2002.

[20] P. Pan, "Tackle How to Identify and Understand High Qualitative AI Data Solutions Will Improve Your Model Performance," in *Interspeech 2022*, Incheon, Korea, 2022.
[Online]. Available: https://www.interspeech2022.org/program/industrytalk.php

[21] S. Radeck-Arneth *et al.*, "Open source German distant speech recognition: Corpus and acoustic model," in *International Conference on Text, Speech, and Dialogue*, 2015, pp. 480–488.

[22] BMBF and Projektträger DLR, *VERBMOBIL II - VM CD21.1 - VM21.1 (ELRA-S0034-30). European Language Resources (ELRA), 1.0, ISLRN 837-421-490- 699-3*. 2004.

[23] V. Pratap, Q. Xu, A. Sriram, G. Synnaeve, and R. Collobert, "MLS: A Large-Scale Multilingual Dataset for Speech Research," *Interspeech 2020*, pp. 2757–2761, 2020.

[24] "LibriVox | free public domain audiobooks." https://librivox.org/ (accessed Dec. 15, 2022).

[25] Bavarian Archive for Speech Signals, "BAS Sl100," 1995. http://hdl.handle.net/11022/1009-0000-0007-E9CF-A (accessed Mar. 28, 2022).

[26] M. Garnerin, S. Rossato, and L. Besacier, "Investigating the impact of gender representation in asr training data: a case study on librispeech," in *3rd Workshop on Gender Bias in Natural Language Processing*, 2021, pp. 86–92.

[27] Y.-A. Chung et al., "W2v-bert: Combining contrastive learning and masked language modeling for self-supervised speech pre-training," *arXiv preprint arXiv:2108.06209*, 2021.

[28] "Deutscher Bundestag - Mediathek." https://www.bundestag.de/mediathek (accessed Dec. 15, 2022).

[29] "Nutzungsbedingungen für das Audio-und Videomaterial des Parlamentsfernsehens." https://www.bundestag.de/mediathek (accessed Dec. 15, 2022).

[30] O. Kuchaiev et al., "NeMo: a toolkit for building AI applications using Neural Modules." arXiv, Sep. 13, 2019. http://arxiv.org/abs/1909.09577

[31] H. Bredin et al., "pyannote.audio: neural building blocks for speaker diarization." arXiv, Nov. 04, 2019. http://arxiv.org/abs/1911.01255

[32] L. Kürzinger, D. Winkelbauer, L. Li, T. Watzel, and G. Rigoll, "CTC-Segmentation of Large Corpora for German End-to-end Speech Recognition," *arXiv:2007.09127*, 2020, doi: 10.1007/978-3-030-60276-5_27.

[33] Z. You, S. Feng, D. Su, and D. Yu, "Speechmoe: Scaling to large acoustic models with dynamic routing mixture of experts," *arXiv preprint arXiv:2105.03036*, 2021.

[34] Z. You, S. Feng, D. Su, and D. Yu, "Speechmoe2: Mixture-of-Experts Model with Improved Routing," in *ICASSP 2022-2022 IEEE International Conference on Acoustics, Speech and Signal Processing (ICASSP)*, 2022, pp. 7217–7221.

[35] Y. Zhang, S. Sun, and L. Ma, "Tiny transducer: A highly-efficient speech recognition model on edge devices," in *ICASSP 2021-2021 IEEE Intl Conf on Acoustics, Speech and Signal Processing (ICASSP)*, 2021, pp. 6024–6028.

[36] X. Shi, P. Zhou, W. Chen, and L. Xie, "Darts-Conformer: Towards Efficient Gradient-Based Neural Architecture Search For End-to-End ASR," *arXiv preprint arXiv:2104.02868*, 2021.

[37] S. Kim et al., "Q-ASR: Integer-only Zero-shot Quantization for Efficient Speech Recognition," *arXiv preprint arXiv:2103.16827*, 2021.

[38] M. Ravanelli et al., "SpeechBrain: A General-Purpose Speech Toolkit." 2021.

[39] E. Bakhturina, V. Lavrukhin, and B. Ginsburg, "NeMo Toolbox for Speech Dataset Construction," *arXiv preprint arXiv:2104.04896*, 2021.

[40] D. Bermuth, A. Poeppel, and W. Reif, "Scribosermo: Fast Speech-to-Text models for German and other Languages," *arXiv preprint arXiv:2110.07982*, 2021.

[41] M. A. Ulasik, M. Hürlimann, F. Germann, E. Gedik, F. Benites, and M. Cieliebak, "CEASR: a corpus for evaluating automatic speech recognition," in *12th Language Resources and Evaluation Conference*, 2020, pp. 6477–6485.

[42] Y. Sinha and I. Siegert, "Improving the Accuracy for Voice-Assistant conversations in German by combining different online ASR-API outputs," *Human Perspectives on Spoken Human-Machine Interaction*, pp. 11–16, 2021.

[43] A. Lopez, A. Liesenfeld, and M. Dingemanse, "Evaluation of Automatic Speech Recognition for Conversational Speech in Dutch, English and German: What Goes Missing?," in *Proceedings of the 18th Conference on Natural Language Processing (KONVENS 2022)*, 2022, pp. 135–143.